%% file: main.tex
\documentclass[journal]{IEEEtran}

\usepackage[belowskip=-6pt,aboveskip=0pt]{caption}
\usepackage{amsmath}
\usepackage{algorithm} 
\usepackage{algpseudocode} 
\usepackage{cite}
\usepackage[pdftex]{graphicx}
\usepackage{gensymb}
\usepackage{color}
\usepackage{soul}
\usepackage[normalem]{ulem} 

\usepackage{etoolbox}
\usepackage{epsfig} 
\usepackage{mathptmx} 
\usepackage{times} 
\usepackage{amsmath} 
\usepackage{amssymb}  
\usepackage{graphicx}
\usepackage{epstopdf}
\usepackage{color}
\usepackage{gensymb}
\usepackage[normalem]{ulem} 
\usepackage{tabularx,ragged2e,booktabs,caption}
\usepackage{subfig}
\usepackage{caption}
\usepackage{soul}
\usepackage{tabularx,ragged2e,booktabs,caption}
\usepackage{textcomp}

\usepackage{setspace}
\usepackage{yhmath}
\usepackage{bm}
\usepackage{makecell}
\graphicspath{ {figure/} }
\input{math/rap2texdefs}

\begin{document}
\title{Towards Safe Control of Continuum Manipulator Using Shielded Multiagent Reinforcement Learning}
\author{Guanglin Ji$^{1\dagger}$, Junyan Yan$^{1\dagger}$, Jingxin Du$^{1\dagger}$, Wanquan Yan$^{1\dagger}$, Jibiao Chen$^{1}$, Yongkang Lu$^{1}$, Juan Rojas$^{1}$,

and Shing Shin Cheng$^{1*}$
\thanks{The research reported in this publication was supported in part by CUHK Direct Grant, CUHK T Stone Robotics Institute (VC Fund 4930807), Innovation and Technology Commission of Hong Kong (ITS/389/18 and ITS/226/19), Research Grants Council (RGC) of Hong Kong (CUHK 24201219), project BME-p7-20 of the Shun Hing Institute of Advanced Engineering, CUHK, and USyd–CUHK Partnership Collaboration Awards. The content is solely the responsibility of the authors and does not represent the official views of the sponsors.}
\thanks{$\dagger$These authors contributed equally to the work.}
\thanks{$^{1}$The authors are with the Department of Mechanical and Automation Engineering, The Chinese University of Hong Kong, Hong Kong.
{\footnotesize $*$Corresponding author, email: sscheng@cuhk.edu.hk}}
}
\markboth{}%
{Shell \MakeLowercase{\textit{et al.}}: Control of Continuum Robot Using Shielded Multiagent Deep Reinforcement Learning}
\maketitle

\begin{abstract}
\textcolor{black}{Continuum robotic manipulators are increasingly adopted in minimal invasive surgery. However, their nonlinear behavior is challenging to model accurately, especially when subject to external interaction, potentially leading to poor control performance.
In this letter, we investigate the feasibility of adopting a model-free multiagent reinforcement learning (RL), namely multiagent deep  Q network (MADQN), to control a 2-degree of freedom (DoF) cable-driven continuum surgical manipulator. The control of the robot is formulated as a one-DoF, one agent problem in the MADQN framework to improve the learning efficiency. Combined with a shielding scheme that enables dynamic variation of the action set boundary, 
MADQN leads to efficient and importantly safer control of the robot. 
{\color{black}Shielded MADQN enabled the robot to perform point and trajectory tracking with submillimeter root mean square errors under external loads, soft obstacles, and rigid collision, which are common interaction scenarios encountered by surgical manipulators. The controller was further proven to be effective in a miniature continuum robot with high structural nonlinearitiy, achieving trajectory tracking with submillimeter accuracy under external payload.}}

\end{abstract}


%
\IEEEpeerreviewmaketitle

\section{Introduction}
\IEEEPARstart Minimally invasive surgery (MIS) involves the use of surgical instruments to access lesion targets in human body through small incision sites. 
MIS offers reduced risks and trauma to the patient, and with faster recovery. Soft continuum robots, which are an emerging category of robots, are expected to play an important role in MIS due to their distal dexterity and structural compliance~\cite{burgner2015continuum}. 
However, the control of these robots often suffers from inaccurate modeling of the flexible behavior due to its structural deformation, interaction with soft tissue, and collision with other instruments, and hysteresis. The significant control error and instability could be fatal in procedures that involve critical structures such as the brain and the heart. 

Various control methods have been developed to improve the motion accuracy of continuum surgical robots. They can be generally categorized into model-based and model-free approaches~\cite{chikhaoui2018control}. 
In terms of the model-based methods, 
piecewise constant curvature assumption is commonly adopted in existing models of continuum robots~\cite{webster2010design}. It allows computationally efficient closed-form solution for many continuum robots in their ideal situations, but environmental interaction and other nonlinear factors are ignored. Cosserat rod model allows external factors to be taken into consideration but are computationally expensive for real-time control~\cite{alqumsan2019robust}. {\color{black}Finite element-based method~\cite{bieze2018finite} has also been developed for continuum robot control.} However, most model-based approaches suffer from the unknown disturbances on the robot in the soft tissue environment. The difficulty in integrating flexible force sensor with continuum surgical robot, due to space, sterilization, and cost, makes it challenging to obtain real-time contact force required in some model-based controllers.

In terms of model-free methods, no prior knowledge about the robot model is needed. Machine learning techniques, such as feedforward neural network, extreme learning machine, and Gaussian process regression~\cite{xu2017data},
have been applied to model the inverse kinematic model and the Jacobian of soft robots. However, these methods may suffer from the fatal inability to address real-time changes in the environment that the robot did not observe during training. Empirical methods and data-driven techniques have been applied to update the model during real-time control. {\color{black}For example, a task space closed-loop optimal controller was developed based on empirically estimated Jacobian~\cite{yip2014model}. In the work by Li et al.~\cite{li2017model}, the Jacobian matrix was estimated using an adaptive Kalman filter that improves convergence and tracking performance.} 
However, the control performance still relies heavily on the accuracy and frequency of the measured data. They may also suffer from singularity issue, estimation lag, and hysteresis-induced errors. 
Hybrid offline and online learning methods have also been proposed. The offline model could be trained using conventional model, finite element analysis, and experimental data on neural networks while online adaptive scheme is developed to update the parameters of locally weighted projection regression, local Gaussian process regression models~\cite{lee2017nonparametric}
or proportional-integral-derivative (PID) controller~\cite{wang2021hybrid}. {\color{black}While online learning methods adapt to environmental interactions, it requires training with a large data set. The need for heurestic parameter tuning also potentially limits the online learning efficiency.}

Reinforcement learning (RL), both model-based and model-free, have recently been investigated for general continuum robot control. A model-based RL was developed to allow computationally efficient closed-loop dynamic control on a soft manipulator by learning the forward dynamics, sampling the control policies, and learning a predictive controller~\cite{thuruthel2018model}. A model-based RL without explicit prior knowledge was implemented for visual-servoing of a continuum robot during simple target tracking~\cite{liu2020distance}. Model-free approaches, such as Q-learning, were implemented on a pneumatic actuated planar soft manipulator~\cite{you2017model,jiang2021hierarchical}. {\color{black}Deep RL, such as deep Q network (DQN)~\cite{mnih2013playing} and deep deterministic policy gradient (DDPG)~\cite{lillicrap2015continuous}, were applied on a soft continuum manipulator with complex nonlinear 3-dimensional (3D) motion.} A Cosserat rod model was used to generate sufficient data to pretrain the network. Open loop and closed loop control were demonstrated to address a range of issues including external loading and workspace discontinuity~\cite{satheeshbabu2019open,satheeshbabu2020continuous}. These model-free approaches however require large datasets for training. Besides, vanilla DQN only allows discrete actions from the action sets, which leads to the inconvenient tradeoff between resolution and efficiency. 
{\color{black}Multiagent RL (MARL)~\cite{bucsoniu2010multi}, which is capable of decentralizing a complex system for increased learning efficiency, was recently introduced on multi-degree of freedom (DoF) rigid robots~\cite{ansari2016multiagent,perrusquia2020multi}.} A MARL with on-policy temporal difference learning was also implemented on a soft arm to perform simultaneous position and stiffness control~\cite{ansari2017multiobjective}. {\color{black}The linear function approximators were used to approximate the Q value, and eligibility trace table was used to record the backward view of temporal difference learning. However, the Q-table and eligibility trace table could become challenging to handle for a large state space.}
{\color{black}In this letter, we investigate the feasibility of using multiagent DQN (MADQN)~\cite{diallo2017learning} algorithm with a shielding strategy~\cite{alshiekh2018safe}, as shown in Algorithm 1, for accurate position control of a continuum surgical manipulator.} Promising results were shown in experiments under various types of external interaction potentially encountered during MIS. The primary contributions are:\par
(a) Constructing the 2-(DoF) continuum manipulator control as a one DoF, one agent problem in the MADQN framework to improve learning efficiency. This is to the best of our knowledge the first attempt in applying MADQN on continuum manipulators.\par
(b) Introducing the shielding scheme in MADQN to permit continuous boundary changes of the action set, effectively removing the discrete action set limitation in the conventional DQN framework. It improves the control performance of the continuum robot control in terms of its targeting accuracy and motion stability. \par
\begin{figure}[t]
	\centering
	\includegraphics[scale=0.22]{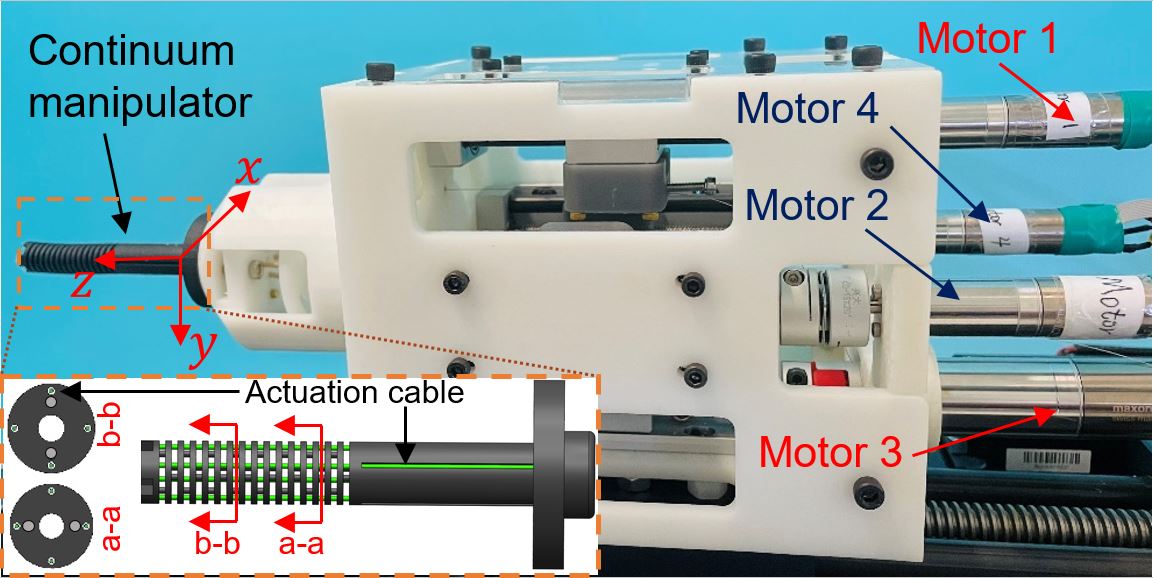}
	\caption{2-DoF continuum surgical manipulator actuated by four motors. Inset: Close-up view of the manipulator showing four cables and the cross sectional views at two different locations along the manipulator.}
	\label{endEffectMotors}
\end{figure}
%
%
\section{Robot Design and Control Architecture}
\subsection{Continuum surgical robot design}
{\color{black}We have designed a 2-DoF continuum surgical manipulator with {\color{black}a 30~mm long flexible segment} and 8~mm diameter~\cite{ding2021towards} to allow active distal manipulation of surgical instruments. It has a central lumen with 3~mm diameter, sufficiently large to accommodate one or more instruments such as a miniature camera, surgical forceps, and suction and irrigation tubing. It consists of a series of 16 disks evenly distributed along the structure with two cylindrical rods connecting every two disks to function as a supporting backbone and the bending joint. Adjacent joints are arranged orthogonally to each other to achieve 2-DoF bending motion. {\color{black}It was 3D printed using stereolithography in the sterilizable high strength nylon}\footnote{ The nylon is a HP3DHR-PA12, Wenext, China.} to allow potential integration with imaging modalities such as magnetic resonance imaging (MRI). It is redundantly driven by four motorized lead screws via four {\color{black}pretensioned} cables with two motors responsible for each DoF as shown in Fig.~\ref{endEffectMotors}. 
Motors 2 and 4 are responsible for the pitch motion (x-z plane) and are controled by agent 1 in our MADQN framework. Motors 1 and 3 are responsible for the yaw motion (y-z plane) and are controled by agent 2.}

\subsection{Markov Decision Process (MDP)}
{\color{black}Most RL agents take advantage of the Markov property of MDPs,}
{\color{black}and the simplest MDP can be represented as a tuple $<S,A,P,R,\gamma> $, where $S$, $A$, $P$, $R$, and $\gamma$ denote the state, action, transition model, reward, and discount factor. The details of MDP can be found in~\cite{sutton2018reinforcement}, and the practical definitions of the task specific parameters in our control setting are defined as follows:}

\noindent $\bullet$
State (\emph{S}): {\color{black}Each state $s \in S$ is determined as $\left(\delta^1,\delta^2 \right)=\left(x_{tar}-x_{tip},\, y_{tar}-y_{tip}\right)$, where $\delta^1$ and $\delta^2$ are the the distance between the robot's tip position and the target position in the pitch DoF and yaw DoF, respectively.} $x_{tip}$ and $y_{tip}$ are the coordinates of the tip while $x_{tar}$ and $y_{tar}$ are the coordinates of the target. The robot is fixed on a platform and its tip has a range of bending angles from 0\degree~to 90\degree. A specific $(x,y)$ coordinate pair in the robot workspace is uniquely matched with a $z$ coordinate. 

\noindent $\bullet$ Action (\emph{A}): {\color{black}There are a pair of motors in each of the two agents in our system. They operate in such a way that if a signal is sent to one motor to pull a cable by a specific displacement, the other motor in the pair will loosen its cable by the same displacement simultaneously.} {\color{black}We define an action set $A = \{-0.2,-0.1,0.1,0.2\}$, shared by each of the two agents. At each iteration, each agent selects an action $a\in A$, according to the current reward, where $a$ represents the cable displacement in mm. It should be noted that $A$ is a predefined action set with no shielding effect added. } 

\noindent $\bullet$ Reward (\emph{R}): {\color{black}The reward function provides a scalar numerical signal as feedback to the agents.} The reward function for agent 1, is defined as below:
{\color{black}\begin{equation}\label{r_x}
	\mu_t^1=
	\begin{cases}
		+1, &\mbox{if } \delta^1_t-\delta^1_{t-1}<0\\
		-1, &\mbox{if } \delta^1_t-\delta^1_{t-1}>0
	\end{cases}
\end{equation}
\begin{equation}\label{R_x}
	r_t^1 = -|\delta^1_t|+\mu_{t}^1
\end{equation}
}
{\color{black}The first term on the right hand side of Eq.~(\ref{R_x}) is the main reward function that ensures the reward value $r^1_t$ is larger for smaller $\delta^1_t$ and it will reach 0 (the highest possible reward value) when the robot tip reaches the target. This will penalize every transition made by the system and force the agent to find the optimal path to the target.  $\mu_t^1$ plays a supporting role to increase the reward if the robot is moving towards the target and reduce the reward if the robot is moving away from the target. The reward function for agent 2 is the same as Eqs. (\ref{r_x}) and (\ref{R_x}), except that superscripts $1$ should be replaced by superscripts $2$.}


\setlength{\textfloatsep}{0pt}
\begin{algorithm}[t]
	{\color{black}
	\caption{\color{black}Shielded MADQN control scheme} \label{TDQN}
	\begin{algorithmic}[1]
	\State Initialize replay buffers $D_{1},D_{2}$ to the capacities $N_{1},N_{2}$, action-value functions $Q_{1},Q_{2}$ with random weight $\theta_{1},\theta_{2}$, and target action-value functions $\hat{Q}_{1},\hat{Q}_{2}$ with random weights $\theta_{1}^-=\theta_{1},\theta_{2}^-=\theta_{2}$
		\For {$episode=1,2,\ldots M$}
		    \State Initialize state {\color{black}$s_t$, $r_t^{1(2)}=0$}
			\For {$t=1,2,\ldots,T$}
	        	\State {\color{black}$a_{t}^1, a_{t}^2 = MADQN.chooseAction$}
	        	\State Implement shielding to obtain safe actions {\color{black}$(a_t^{1s},a_t^{2s})$}
				\State Execute joint action {\color{black}$(a_t^{1s},a_t^{2s})$} on the motor and observe reward $r_{t}^1,r_{t}^2$, and new state $s_{t+1}$
				\State Store transition $({\color{black}s_{t},a_{t}^1,r_{t}^1, s_{t+1})}$ and {\color{black}$(s_{t},a_{t}^2,r_{t}^2,s_{t+1})$} in $D_{1},D_{2}$
				\State Sample random minibatch of transitions {\color{black}$(s_{j},a_{j}^{1(2)}, r_{j}^{1(2)},s_{j+1})$ from $D_{1},D_{2}$}
				\State{\color{black}Calculate temporal difference target $y_j^{1(2)}$}
				\State {\color{black}Perform a gradient descent step on $(y_{j}^{1(2)}-Q(S_j,a_j^{1(2)};\theta_{1(2)}))^2$ according to Eq.~(\ref{loss})} 
				\State Reset $\theta_{1}^-=\theta_{1}$ and $\theta_{2}^-=\theta_{2}$ every C steps
			\EndFor
		\EndFor
	\end{algorithmic} 
	}
\end{algorithm}

\subsection{Formulation of Multiagent Deep Q Network (MADQN) with shielding}

\subsubsection{Multiagent Deep Q Network}
{\color{black}The MADQN framework is a multiagent extension of the original Q learning algorithm and subsequently the novel DQN. It was introduced to explore how agents can behave cooperatively to maximize a cumulative reward in a shared environment~\cite{diallo2017learning}. In this work, we propose to apply it on the 2-DoF continuum manipulator with each agent representing the actuator pairs of each DoF.} {\color{black}The output of the MADQN neural network (NN) can then be halved compared to the single agent NN, allowing a sharp decrease in the training time.}

{\color{black}Using multiple agents can theoretically improve the robustness of the robotic system but will also hurt the system's stability, as the action taken by one agent will affect the states of other agents.}
In our {\color{black}learning} architecture, each agent is designed to train separately. {\color{black}The policy of agent 1 is learned only based on the error in the $x$ direction (see Fig.~\ref{endEffectMotors}). The error caused by agent 2 in the $y$ direction will not affect the performance of agent 1 due to the small correlation between the two DoFs in the continuum robot. As a result, the instability problem introduced by the use of multiagent is not a concern in our application.} {\color{black}Note that the termination condition in the control framework is defined in terms of the Euclidean distance between the robot tip position and the target position in the task space, which ensures the states are shared by the two agents. This setting would facilitate the robot tracking error convergence.}

Figure 2 illustrates the basic control diagram of MADQN. 
{\color{black}At each time-step, agents will choose corresponding motor actions according to the current errors along different directions}. Then the joint action signal of the two agents will be sent to the motors to actuate the robot. After that, a new state and respective reward for each agent will be observed. 

\begin{figure}[h!]
\centering
\includegraphics[scale=0.26]{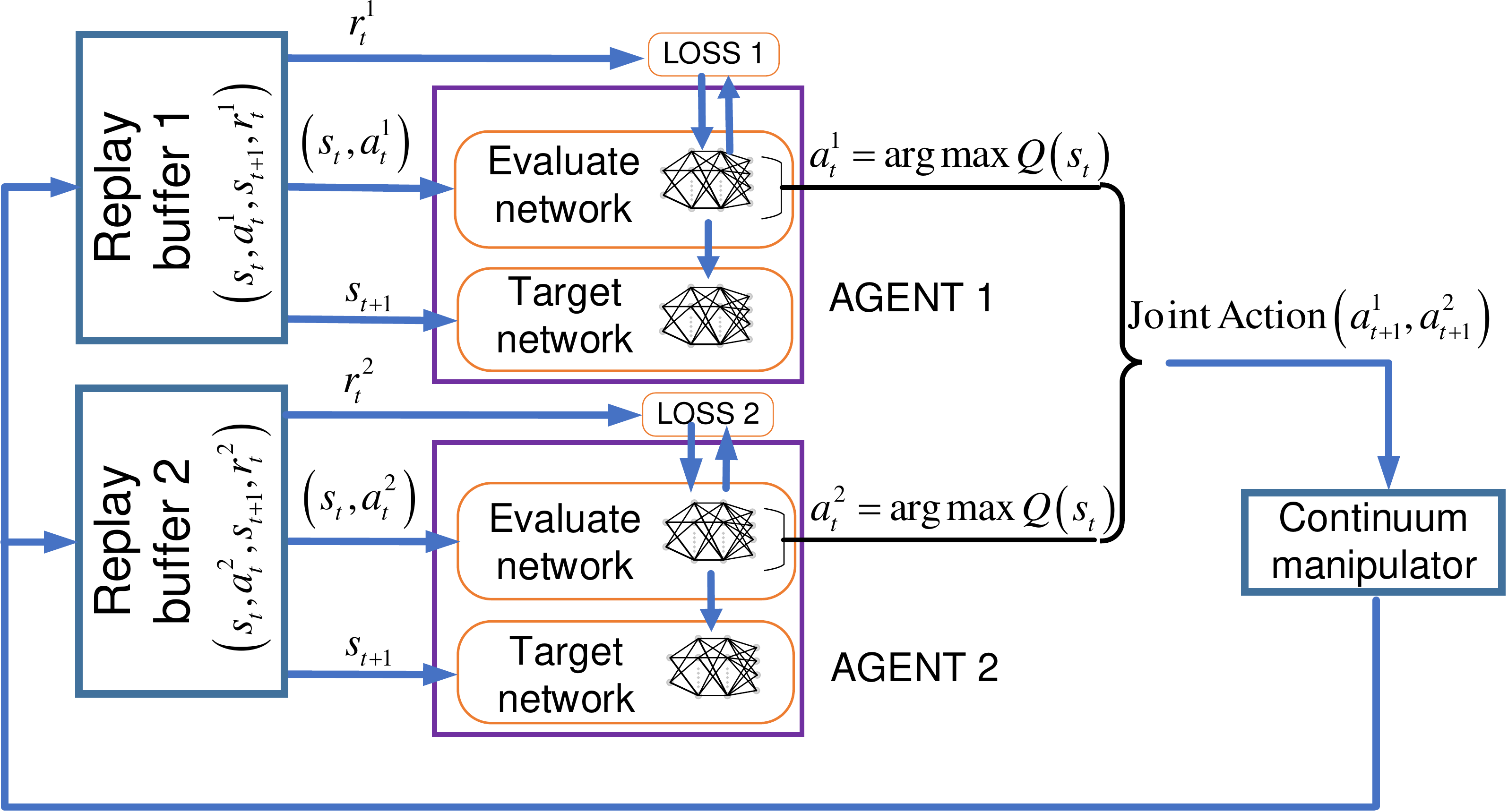}
\caption{{\color{black}MADQN framework consisting of evaluate network and target network for each agent.}}
\label{fig:RL Framework}
\end{figure}

In the MADQN algorithm, the Q value is approximated by NNs similar to DQN. There are two neural networks, namely evaluate network and target network, in each agent. The evaluate network is used to evaluate the current action value function {\color{black}$Q (s_{j},a_{j};\theta)$}, and $\theta$ is the vector containing weights of the network. The output of the target network $\hat{Q}(s_{j+1},a_{j+1};\theta^{-})$ is used to calculate the target of the $j$-th iteration:
{\color{black}\begin{equation}
	y_{j}=\mathbb{E}_{s_{j+1}\sim\kappa}\left[r_{j}+\gamma \max\limits_{a_{j+1}}\hat{Q}(s_{j+1}, a_{j+1}; \theta^{-}_{j-1})|s_j,a_j\right]
\end{equation}}
where $\kappa$ is a state emulator. The neural networks are trained by minimizing a sequence of loss functions
{\color{black}\begin{equation}\label{loss}
	L_j\left(\theta_j,\theta^-_j\right)=\mathbb{E}_{s,a\sim\rho}\left[\left(y_j-Q\left(s_j,a_j;\theta^-\right)\right)^2\right]
\end{equation}}
where $\rho\left(s,a\right)$ is a joint probability distribution of state $s$ and action $a$.
An epsilon-greedy policy is used to balance the exploration and exploitation problem during learning. The parameter $\epsilon$ indicates the degree of greed. 
In MADQN, a replay buffer is used to store the transitions and continue the learning process during testing, leading to its capability to adapt to changes in the environment that is not available during the off-policy training. \par

\subsubsection{MADQN with shielding}
\begin{figure}[h!]
\centering
\includegraphics[scale=0.36]{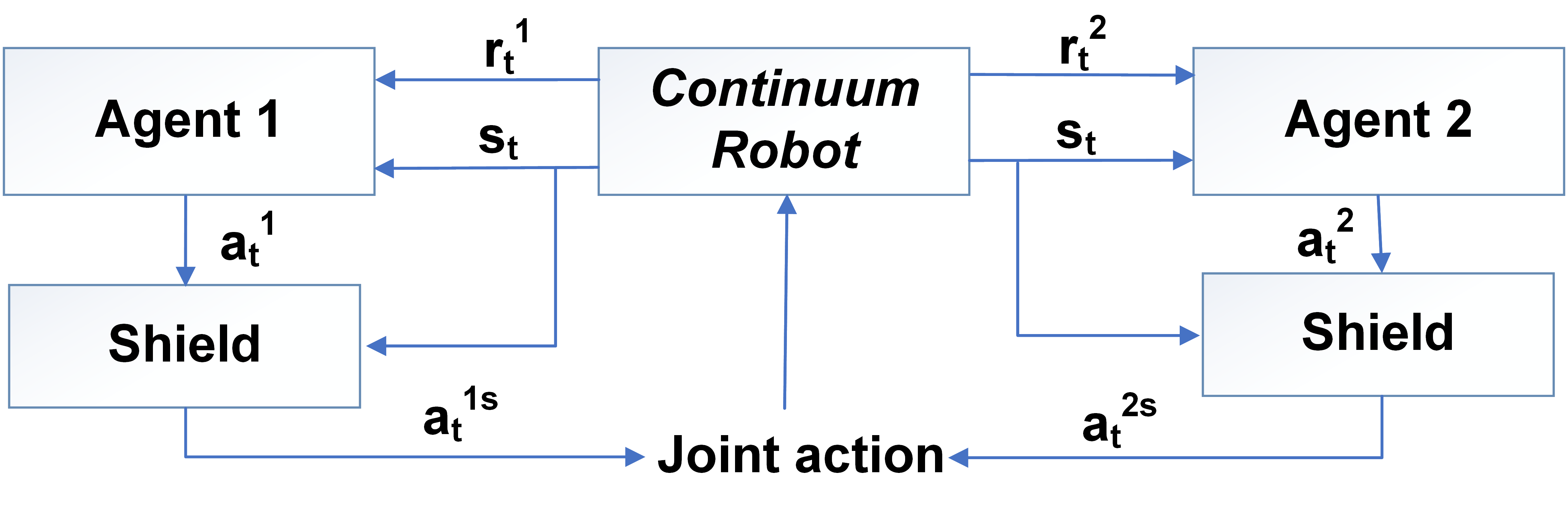}
\caption{{\color{black}Concept illustration of the MADQN framework with a shielding scheme}}
\label{Shielding TDQN Framework}
\end{figure}

When applying RL-based control policies on a continuum surgical manipulator, improving safety concern must be the priority to ensure accurate point and trajectory tracking ability and improve motion stability. The actions must be sufficiently {\color{black}delicate and accurate}, especially when the robot tip is close to the target. However, the tracking performance could be inefficient, especially when the robot begins at a relatively distance away from the target. {\color{black}Therefore, we propose a shielding scheme to allow the actions to be adaptive to the Euclidean distance between the robot tip and the target.} 
{\color{black}It functions by adopting a linearized saturation function that makes the action set boundary linearly proportional to $\delta_t$.} This ensures that robot tip oscillation or any large undesired state change could be avoided when the robot is very close to the target. 
{\color{black}The shield is placed after MADQN, as shown in Fig.~\ref{Shielding TDQN Framework}. The current state and action chosen by each agent are provided to the corresponding shield block before a safer action, $a^{1(2)s}_t$}, adjusted by the shielding scheme, becomes the actual output. 


{\color{black}
\begin{algorithm}[t]
	{\color{black}
	\caption{Shielding algorithm}
     \label{algorithm2}
	\hspace*{\algorithmicindent} \textbf{Input: $s_t$, {\color{black}$a_t$}} \\
    \hspace*{\algorithmicindent} \textbf{Output: {\color{black}$a_t^{s}$}}
	\begin{algorithmic}[1]
	\State $A=[-a_{boundary},...,a_{boundary}]$
	\State Each action set has four actions uniformly distributed between $-a_{boundary}$ to $a_{boundary}$.
	\If{$|\delta|>K$}
	    \State ${\color{black}A^{s}}=[-a_{boundary},...,a_{boundary}]$
	\Else
	    \State $a^s_{boundary}=\beta|\delta|$
	    \State $A^{s}=[-a^s_{boundary},...,a^s_{boundary}]$
    \EndIf
    \State Find original $a_{index}$ from $A$
    \State $a_t^{s}=A^{s}(a_{index})$
	\end{algorithmic} 
	}
\end{algorithm}
}\par
The complete shielding algorithm is shown in {\color{black}Algorithm~\ref{algorithm2}}. {\color{black}$A^s$, $a_{boundary}$, $a^s_{boundary}$, and $a_{index}$ denote the safe action set, the boundary action element of the original action set A, the boundary action element of the safe action set $A^s$. $K$ is the threshold distance that triggers the initiation of the shielding algorithm while $\beta$ is an empirical parameter proportional to the gradient of the adopted linearized saturation function.} {\color{black}In our experiments, $K$ was set to 2~mm after multiple trials and errors to keep a balance between the efficiency and motion smoothness. If $K$ is too large, more steps are needed to arrive to the target position as the actively adjusted safe actions are smaller than those in the original action set. If $K$ is too small, there is serious oscillation when the robot tip is close to the target position. $\beta$ was set to 0.1 to make sure there was not a large difference between the original action values and the safe action values. The minimum $A^s_{boundary}$ was maintained at 0.05~mm to make sure there is always obvious change in the positioning error at each step. When the Eucledian distance between the robot tip and the target position is less than 0.25~mm,  the robot tip is considered to have arrived at the target position and the episode is terminated.}

\section{Simulation}
The MADQN algorithm was first tested in simulation to evaluate its feasibility as a position controller for the continuum surgical robot. {\color{black}A constant curvature model~\cite{webster2010design} was used to simulate the behavior of the continuum manipulator which was designed to have the same geometric dimensions as those described in Section II(A).}\par

{\color{black}During the off-policy training of the MADQN algorithm, the 2-DoF manipulator was commanded to perform a single target point (10 mm, -10 mm) tracking with no shielding. The target point could be any point inside the workspace of the robot. After every episode, the manipulator would be reset to the original point from its last state. It is worth noting that the robot does not have to be trained on the full workspace to learn to move towards the target due to the states being the relative distance {$\delta^1,\delta^2$} instead of the absolute position ($x_{tip},y_{tip}$). Each agent only needs to learn to take an action, in particular positive or negative action, to maximize the reward.}


For the MADQN simulation, we use a fully connected neural network (FCN) for each agent with {\color{black}two input states ($\delta^1,\delta^2$) and four output Q values for the four actions} (see Section II(B)). The FCN uses three hidden layers, each with 180 nodes and followed by ReLu activations. The evaluate network and target network share the same network structure. As for exploration, a greedy threshold begins at 1.0 and decreases to 0.01 by multiplying it with an $\epsilon$ decay rate of 0.9995 during its training. A gradient descent temporal difference error (TD error) step is optimized by the Adam optimizer with a learning rate of 0.0001. Each episode consists of 200 iteration steps. The termination condition is triggered after those 200 steps or when the state error is less than 0.25~mm for 10 iteration steps. The MADQN algorithm was trained for around 80 episodes before being evaluated for trajectory tracking. The hyper parameters needed are shown in Table 1.\par

\begin{figure}[h!]
\centering
\includegraphics[scale=0.35]{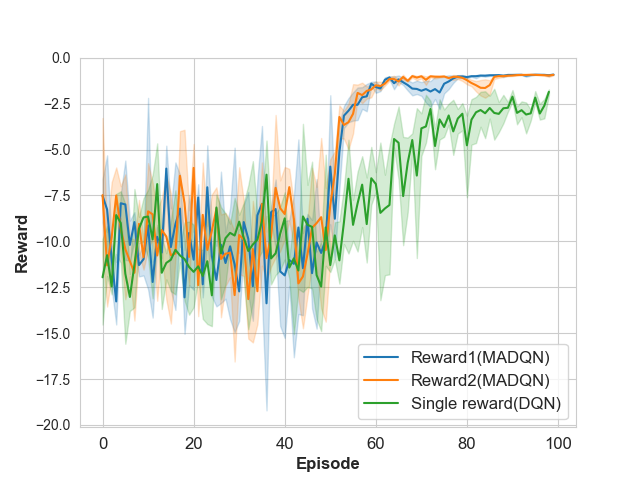}
\caption{{\color{black}Reward comparison between the single-agent DQN and MADQN. The blue and orange lines indicate reward 1 and reward 2 for agent 1 and agent 2, respectively in MADQN while the green line indicates the reward in single agent DQN. The shaded area presents the standard deviation.}}
\label{fig:Accumulated Reward}
\end{figure}

{\color{black}In the single-agent DQN setting, the same neural network was used, but the number of nodes was increased to 240 to facilitate more efficient convergence. }
\begin {table}[t]
\centering
\caption {Hyper parameters}
\begin{tabular}{|c|c|} 
\hline
Discount Factor $\gamma$ &0.95\\
\hline
Replay Buffer Size &10000\\
\hline
Learning Rate $\alpha$ &0.0001 \\
\hline
Batch Size &100\\
\hline
Max Episodes &150\\
\hline
Initial $\epsilon$ &1.0\\
\hline
$\epsilon$ decay rate &0.9995\\
\hline
Target Network Update Frequency &200\\
\hline
\end{tabular}
\end {table}
In Fig.~\ref{fig:Accumulated Reward}, we compare the {\color{green}reward} during training of both the single-agent DQN and MADQN algorithms.
The training time of MADQN is significantly shorter than the single-agent DQN. In MADQN, training stabilized around episodes 60-70, while {\color{black}the single-agent DQN shows unsatisfactory performance even at 100 episodes.  The reward function of the single agent also oscillates much more than that for each of the two agents of MADQN with larger standard deviation, as shown in Fig.~\ref{fig:Accumulated Reward}.} {\color{black}It has also been shown in~\cite{satheeshbabu2019open} that more than 5000 training episodes were required to allow convergence using single agent DQN.} 
\par

\section{Results and Discussion}
\label{sec:experiments}
\subsection{Experimental platform and system integration}
\begin{figure}[h!]
	\centering
	\includegraphics[width=0.65\linewidth]{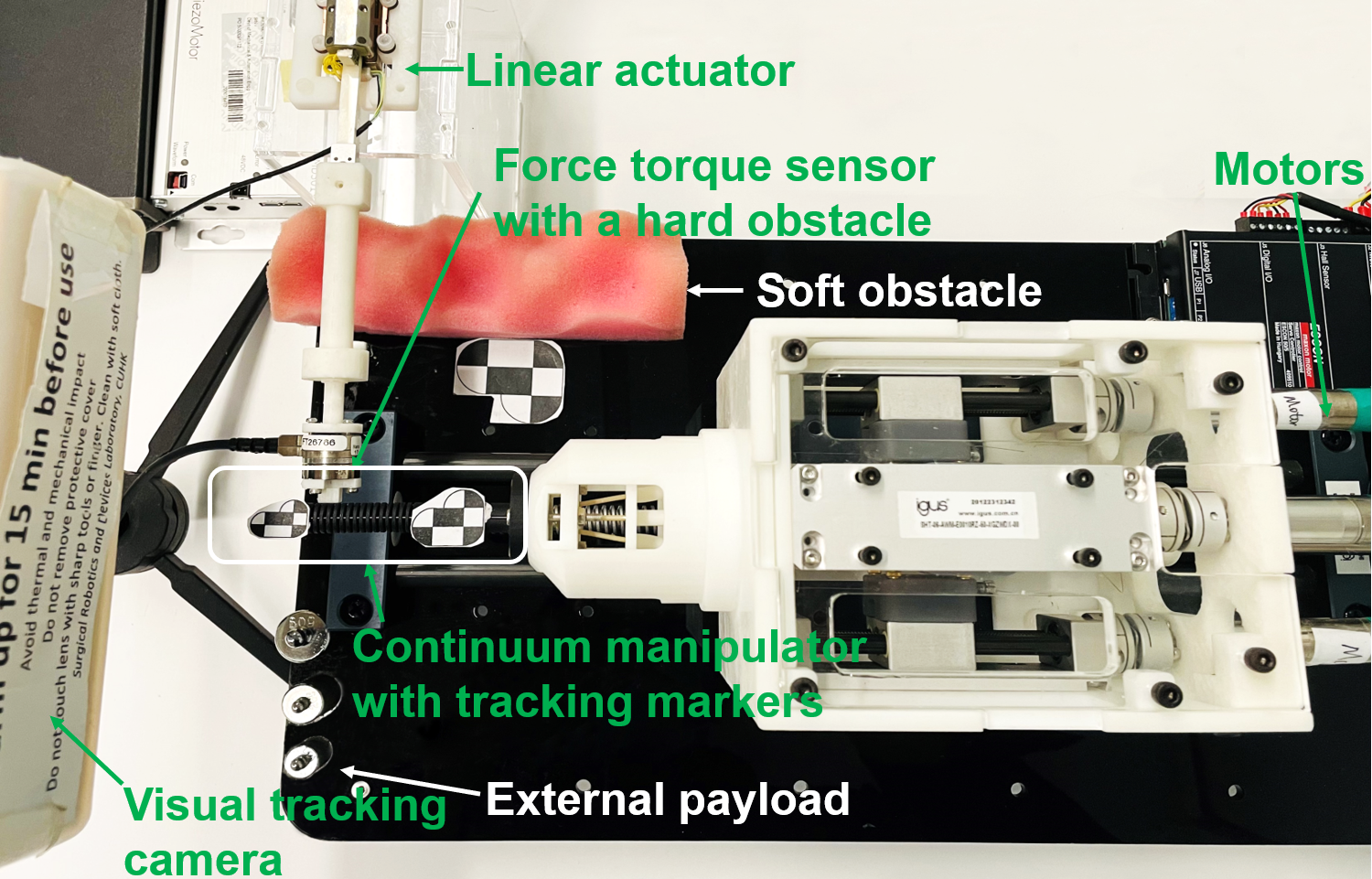}
	\caption{Experimental setup to evaluate the continuum manipulator performance under MADQN. Various components used to simulate the different environmental constraints can be seen, including the external payloads, the soft obstacle, and the hard obstacle actuated by a linear actuator to collide with the manipulator.}
	\label{setup}
\end{figure}

Our continuum surgical manipulator comprises of two computers, a 6-DoF visual tracking camera\footnote{The tracking camera is a MicronTracker, Claron Technology, Canada}, and four DC motors with embedded incremental encoders and motor controllers\footnote{We use the DCX 14 and ESCON 36/2, Maxon Inc. from Switzerland}. The camera measured the tip position of the continuum manipulator. We also used closed-loop feedback control with continuous positional data. Each actuation cable was connected to an adapter on a lead screw module to allow for high resolution cable displacement. Two fiducial markers were attached to the proximal and distal end of the continuum manipulator, which allowed us to measure the bending angle and tip position in 3D space. A data acquisition system\footnote{The data acquisition system used is a Model 826, Sensoray, from the USA} collected all the sensing data, whilst the the control program was implemented in Simulink (Mathworks, USA) and Python 3. The communication between the server and client was realized by Transmission Control Protocol/Internet Protocol (TCP/IP). The server received action signals, delivered the corresponding signals to control the motors, and read and sent sensor data to the client computer. The client computer received the sensor data as input to the MADQN algorithm which used TensorFlow 2.4.0. The data was processed and the best action of the current policy was chosen and sent to the server. 

\subsection{Off-policy learning performance comparison: Single-agent DQN and MADQN}
\label{off-policy}
\begin{figure}[htbp!]
	\centering
	\includegraphics[scale=0.55]{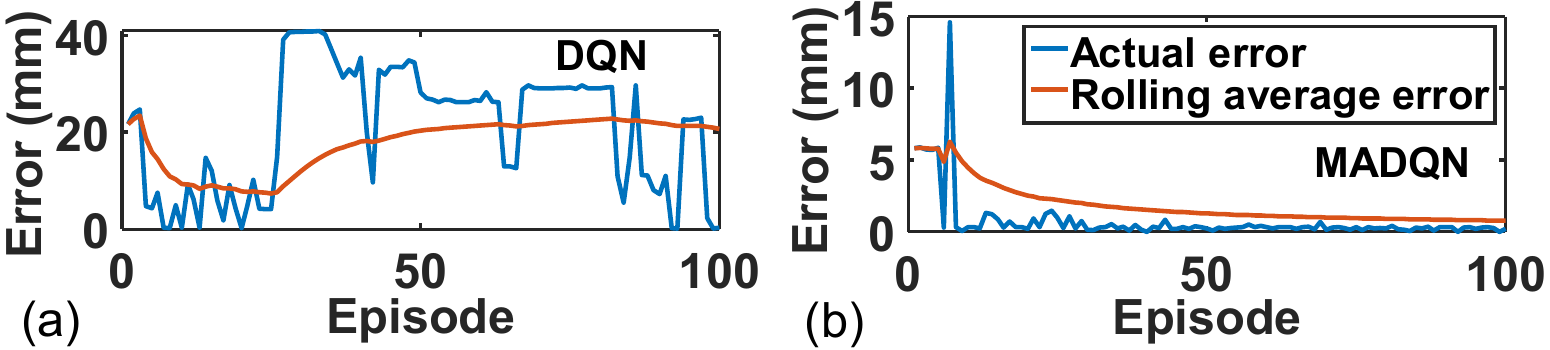}
	\caption{{\color{black}Point tracking error comparison under single-agent DQN and MADQN during off-policy training. The tracking target was (x=10~mm,y=10~mm) in task space coordinates.}}
	\label{ex_15}
\end{figure}

{\color{black}The physical system was trained using off-policy batch learning before performing the actual control task. During this offline learning process, the manipulator would learn to reach a desired target position in free space. Data collection was done on the physical robot instead of using model-based simulation to minimize the errors from model assumptions~\cite{thuruthel2018model}. Similar to the process during the simulation, we trained the system with both the single-agent DQN and MADQN to compare their tracking performance and training time.} 
{\color{black}In terms of the tracking performance, {\color{green}Fig.~\ref{ex_15}} shows that single-agent DQN could not converge within 100 episodes in contrast to MADQN. Furthermore, the rolling average tracking error for MADQN gradually decreased to around 0.25~mm after 100 episodes with RMS errors being 2.23~mm, 0.35~mm, and 0.31~mm after 25 episodes, 50 episodes, and 75 episodes, respectively (red line in Fig. 6(b)).  
In terms of the training time, single-agent DQN used 150~minutes to train for 100 episodes while MADQN used a third of the time (50~minutes). {\color{black}By simplifying the robot control into one DoF-one agent problem, we reduced the dimension of the neural network output by one half and were able to achieve greater training time saving.} As a result, MADQN converged faster and used significantly less iteration numbers per episode. The higher learning efficiency of MADQN in our continuum robot could potentially facilitate its deployment in practical scenarios.}
{\color{black}The following evaluation of the robot in Subsection IV(C)-IV(G) were performed after the robot underwent the off-policy learning in free space with no external payload for 100 episodes of single point tracking of (x=10~mm, y= -10~mm).}

\subsection{Trajectory tracking with and without shielding}
\begin{figure}[!ht]
	\centering
	\includegraphics[scale=0.47]{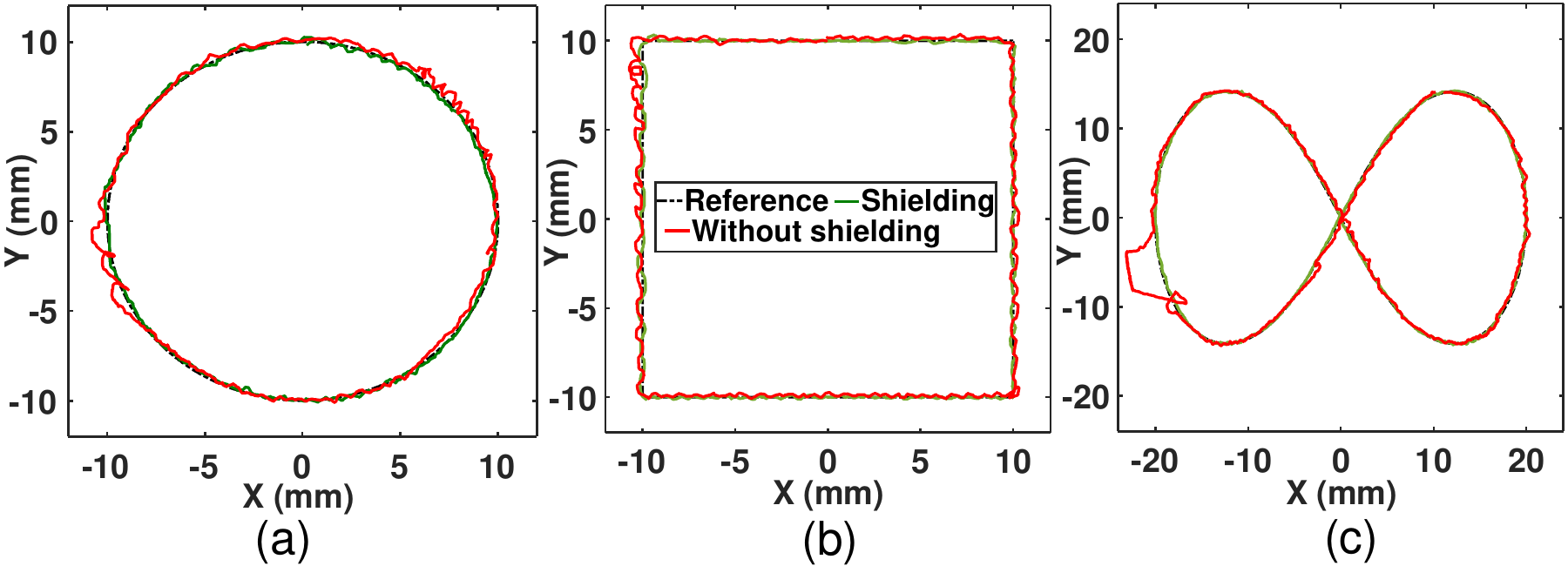}
	\caption{{\color{black}Trajectory tracking results (with and without shielding scheme) for 3 shaped trajectories: (a) Circle, (b) Square, and (c) Infinity. Note that a 20 Hz tracking frequency was used.}}
	\label{ex_2}
\end{figure}
The continuum manipulator was commanded to track three different reference trajectory patterns (circle, square, and an infinity shape) in a free environment under MADQN. {\color{black}It should be noted that the manipulator bends in a hemispherical 3D workspace with each x-y position uniquely matched with a z-coordinate, and the reference trajectories are actually 2D projections on the x-y plane.} Here we evaluate the effect of the shielding scheme. The RMS tracking errors of the three trajectories without shielding were 0.68~mm, 0.54~mm, and 0.71~mm. The corresponding maximum tracking errors are 1.52~mm, 0.83~mm, and 4.79~mm, respectively. When shielding was activated, the RMS tracking errors were reduced to 0.31~mm, 0.28~mm, and 0.24~mm respectively and their corresponding maximum tracking errors were 0.38~mm, 0.39~mm, and 0.36~mm, respectively. {\color{black}While MADQN allowed relatively acceptable trajectory tracking of different shapes without shielding, as seen in Fig.~7, shielding led to significant RMS error reduction with values of 54.4\%, 48.1\%, and 66.2\% for the three shapes, respectively.}
{\color{black}Shielding improved the accuracy as it constrained the motor movement in real-time based on the actual tracking condition. This dampened overshoot and oscillation amplitude, allowing smoother trajectories. Shielding also helped decrease the babbling motion during initialization. All these helped to ensure smoother and more stable robot motion, which would be of high significance considering the confined space in the human body (e.g. heart chamber, oral cavity, brain, and abdominal space) that the robot would be operating in.}

\subsection{Safe control evaluation 1: External payload}
\label{payload}
\begin{figure}[!ht]
	\centering
	\includegraphics[scale=0.5]{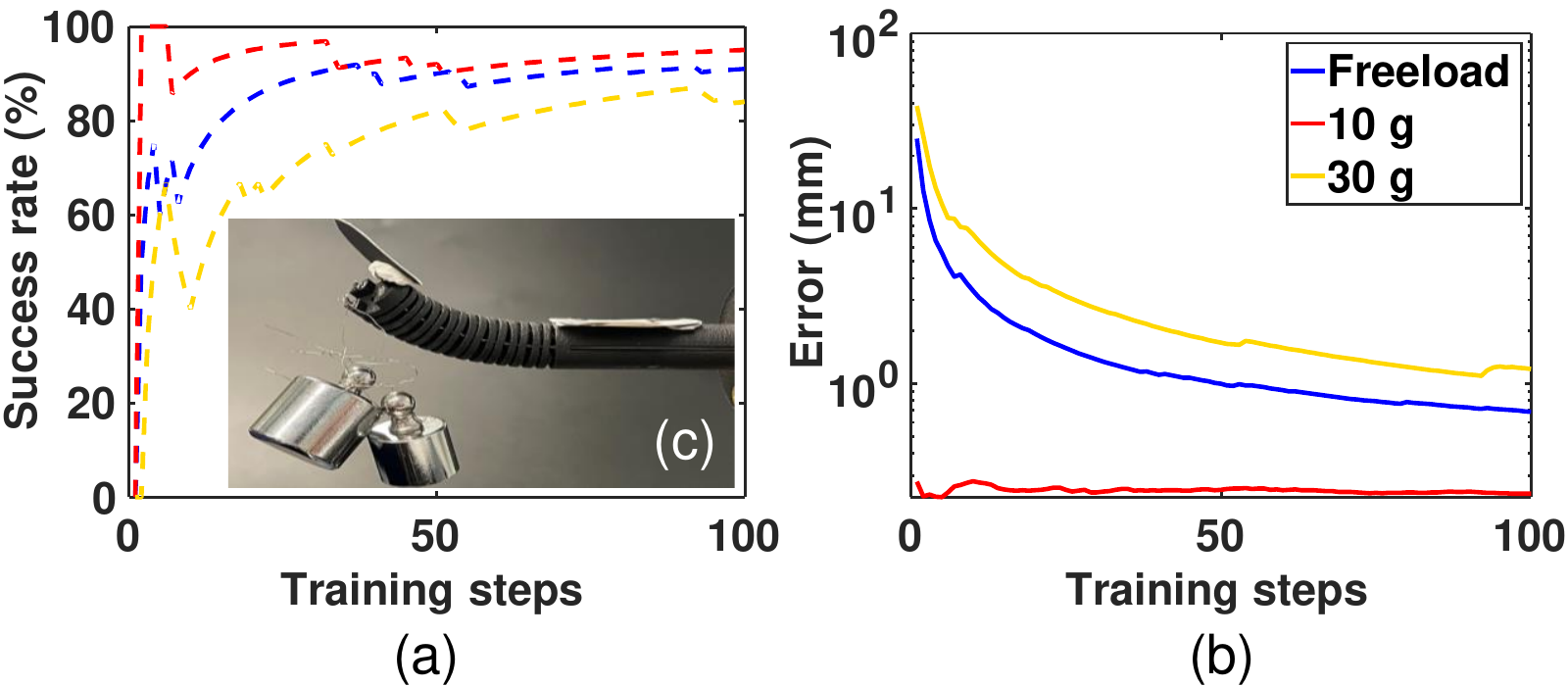}
	\caption{{\color{black}Point-tracking result under different external payloads: (a) Success rate and (b) performance defined in terms of rolling average tracking error.}}
	\label{ex_3}
\end{figure}
\begin{figure}[!ht]
	\centering
	\includegraphics[scale=0.45]{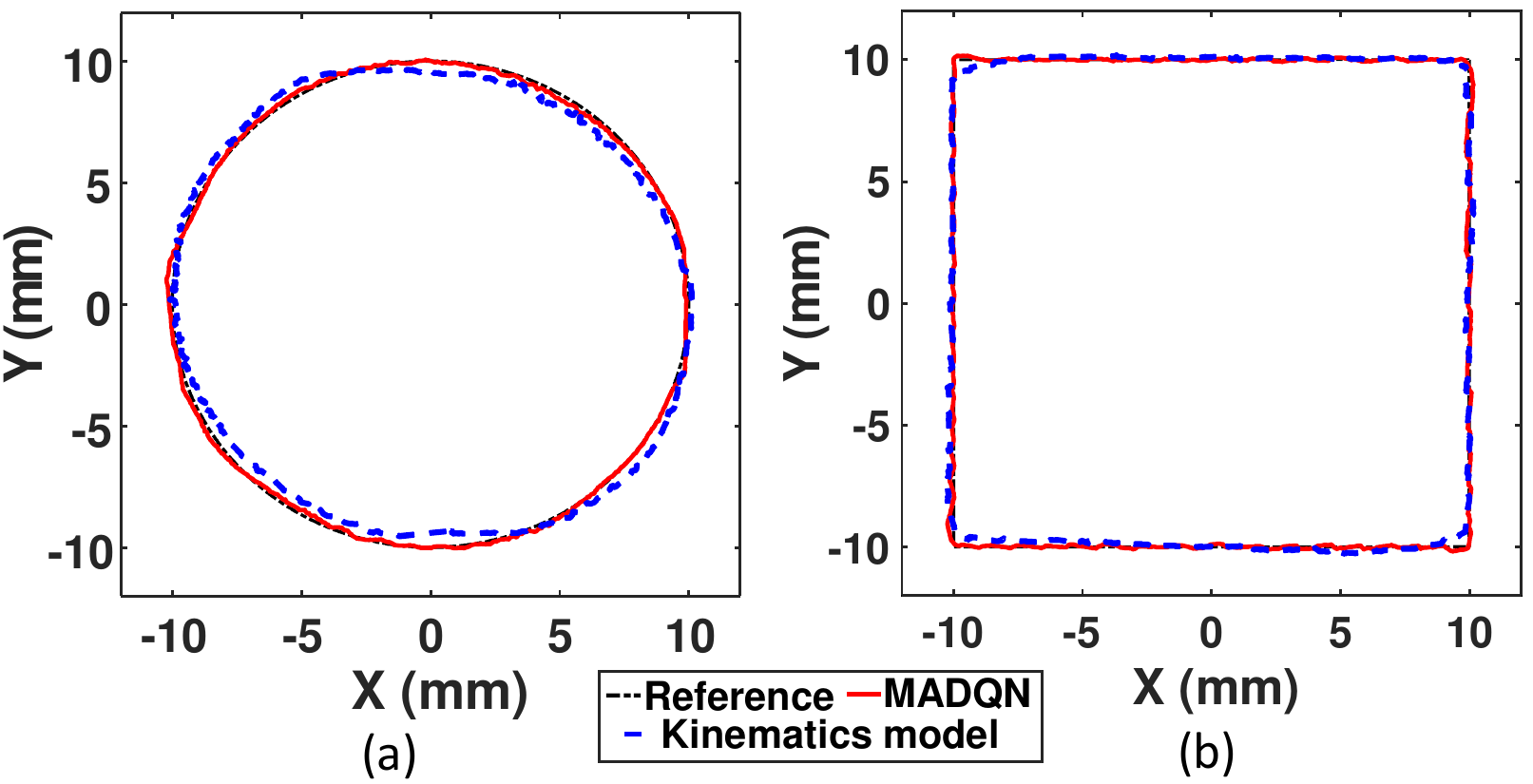}
	\caption{Trajectories tracking results under 30~g external payloads: (a) Circle trajectory and (b) Square trajectory.}
	\label{ex_3b}
\end{figure}
{\color{black}In order to simulate the scenarios where the continuum robot performs tissue sample manipulation or carries instruments such as endoscopic camera or forceps at its tip, we hung different weights (0~g, 10~g, and 30~g) at the manipulator tip before commanding it to track a single point (x=10~mm,y=8~mm) and two 2D projected trajectories (i.e. circle and square). To further evaluate the performance of the proposed MADQN controller, 
a kinematics model-based optimal controller (KMOC)~\cite{sefati2020surgical,ding2021towards} was also implemented during the shape trajectory tracking to serve as a performance comparison to the model-free MADQN. Since the weight of the manipulator was only 8~g, the latter two payloads represented hefty loads for the manipulator at 1.25 and 3.75 times its weight. In fact, the manipulator was deformed upon the addition of the weights. 

During the point tracking experiments under MADQN, success was defined as reaching the target position within a threshold of 0.35 mm. Fig.~\ref{ex_3}(a) illustrates the success rate as a function of the online training steps. Fig.~\ref{ex_3}(b) subsequently shows the improvement with time in the tracking performance in terms of the rolling average error, as the robot kept repeating the step tracking action. The RMS tracking errors under free load, 10~g, and 30~g are 0.22~mm, 0.20~mm, and 0.31~mm, respectively. It should be noted that the robot was trained without any payload as mentioned in Section \ref{off-policy}. The online learning feature of MADQN enabled relatively fast learning speed and low error even with large external payloads. 
During the shape trajectory tracking (only 30~g external payload was used), as shown in Fig.~\ref{ex_3b}, the performance of MADQN (RSME=0.32~mm for the circle trajectory and RSME=0.29~mm for the square trajectory) was better than KMOC (RSME=0.68~mm for the circle trajectory and RSME=0.53~mm for the square trajectory). This is due to the inaccuracy of the analytical Jacobian model developed under the constant curvature assumption in capturing the robot motion under a large external payload. 
} 

\subsection{Safe control evaluation 2: Soft obstacle}
\begin{figure}[!h]
	\centering
	\includegraphics[scale=0.45]{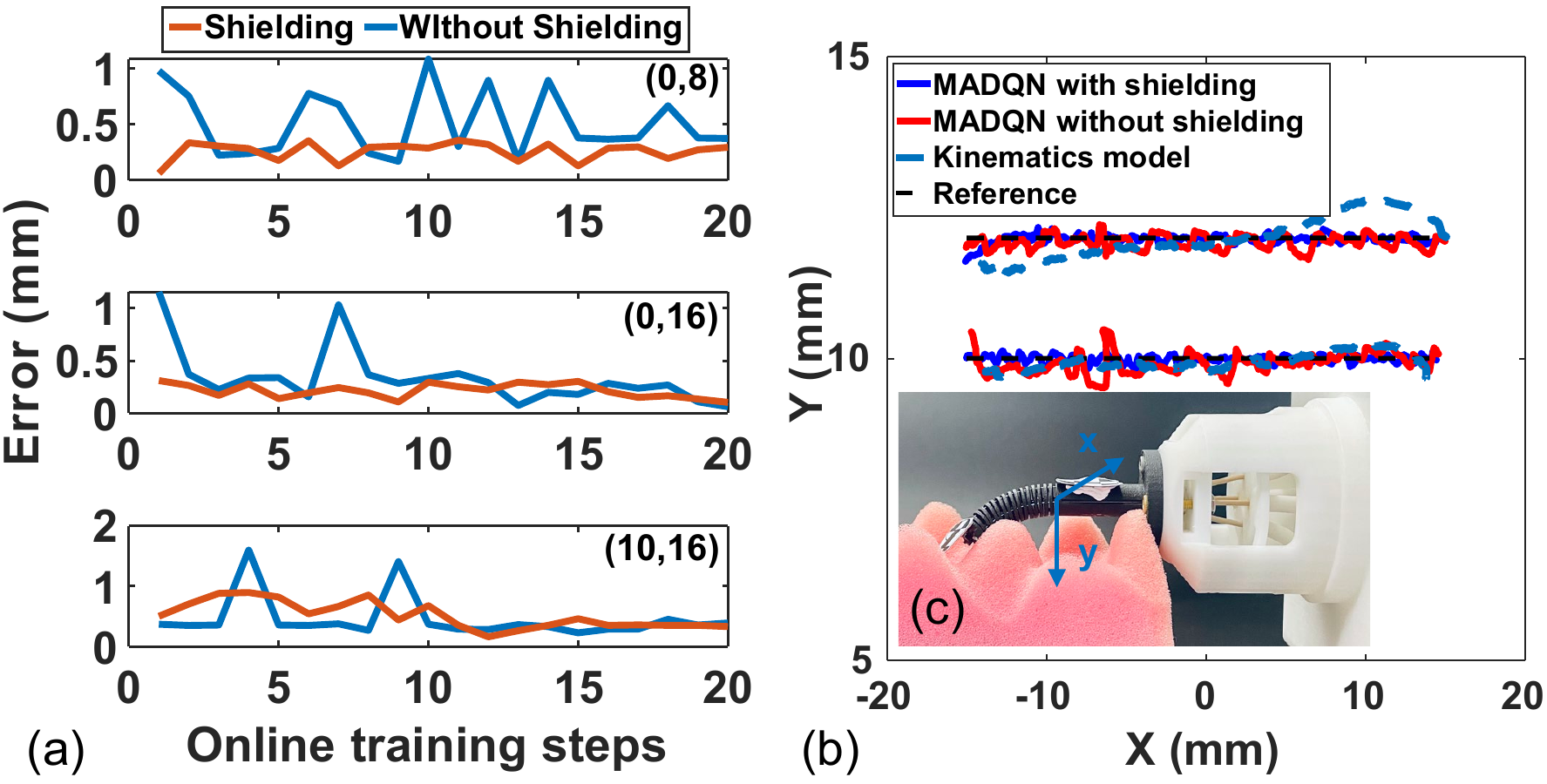}
	\caption{{\color{black}}Point and trajectory tracking results under a soft obstacle constraint. The soft obstacle was in contact with the continuum manipulator throughout its trajectory (a) Point tracking results of three targets with 2D and 3D motion. (b) Trajectory tracking results of two predefined straight lines. Tracking started at (-15,10)~mm and ended at (15,12)~mm.}
	\label{ex_4}
\end{figure}
{\color{black}An "egg crate" foam (see Fig.~10(c)) was placed under the continuum manipulator to serve as an environmental constraint that resembles the soft tissue environment in the gastrointestinal wall and inner linings of other organs.} The goal of this study is to investigate if the online learning capability of the MADQN can improve the tracking performance in the presence of soft perturbation. 
The manipulator used MADQN with and without shielding to perform point tracking to several locations that required either planar (1-DoF) or 3D (2-DoF) manipulator bending. After reaching a target location, the robot would move back to its home (straight) configuration. Three different target points [(0,8)~mm,(0,16)~mm,(10,16)~mm] were selected to ensure the robot was in contact and pressed the foam with varying degree of forces. Each point was tracked 20 times.
The results for these experiments are shown in Fig.~10. Focusing on the shielded MADQN performance, the maximum tracking error of 0.90~mm occurred at the beginning of the experiment but consistently declined with the number of training steps during the 3D bending motion to reach the target (x=10 mm,y=16 mm). As for 2D targets, (x=0 mm,y=8 mm) and (x=0 mm,y=16 mm), tracking errors were steady throughout the 20 steps and produced RMS errors of 0.21~mm and 0.20~mm, respectively. 
Generally, shielding offers significant improvement in the performance under soft disturbance.
{\color{black}
Besides point tracking, we also commanded the manipulator to follow straight-line trajectories that crossed the peaks and valleys of the foam at y=10~mm and y=12~mm under the control of MADQN (with and without shielding) and KMOC. As shown in Fig.~10(b), MADQN with shielding achieved the best performance with RMS errors of 0.31~mm for y=10~mm and 0.35~mm for y=12~mm, respectively, again confirming the superiority of MADQN over the kinematics model-based control. These values are also safely within the acceptable precision range requirement of most surgeries~\cite{shi2016shape}.}
\subsection{Safe control evaluation 3: Rigid collision}
\begin{figure}[!htb]
	\centering
	\includegraphics[scale=0.5]{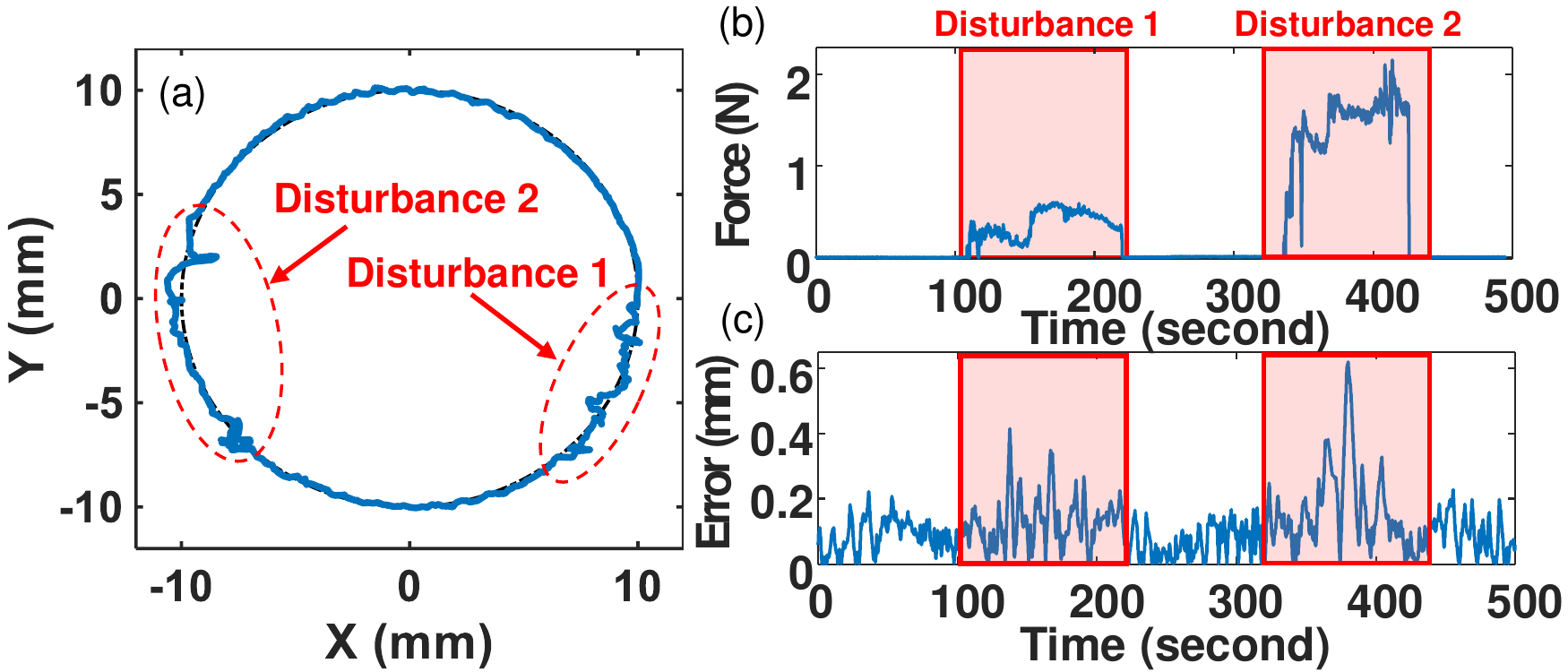}
	\caption{Trajectory tracking results under rigid collision: (a) Tracking results under MADQN with shielding where collision was applied at two periods during the trajectory following. (b) and (c) show the contact force and tracking error against time.}
	\label{ex_5}
\end{figure}

{\color{black}This experiment intends to demonstrate how the proposed control algorithm can help the continuum manipulator accommodate and recover from the influence of external disturbances such as unconscious collision with rigid obstacles during its dynamic motion. Intracorporeal collision is a common scenario when two or more surgical instruments are manipulated in close proximity~\cite{tang2012ergonomics}, especially in single port access surgery (SPAS), causing uncontrolled instrument motion and compromising patient safety. In this experiment, we simulated the rigid collision by pushing a plastic rod using a linear actuator (LT-40, PiezoMotor, Sweden) against the {\color{black} middle section of the }continuum manipulator, as seen in Fig.~\ref{setup}, during its motion along a 2D projected circular trajectory and maintaining the contact for about 100~s.}
For this experiment, a force and torque sensor \footnote{The FT sensor is a Nano17, ATI Industrial Automation, US.} was attached at the tip of the rod to measure the contact force. A plastic rod was displaced until it collided with the manipulator, and it maintains its position for some time before receding. The rod was commanded to repeat this contact twice at two different locations while the continuum robot performed a circular trajectory. Fig.~\ref{ex_5} illustrates the disturbance locations, durations, and tracking errors. The collisions occurred at the 100$^{th}$ and 330$^{th}$ second in the experiment and applied a 2N force. The MADQN algorithm with shielding allowed the closed-loop controller to accommodate for the disturbance resulting in a maximum tracking error of 0.45~mm and 0.62~mm for each of the two disturbances. {\color{black}This result indicates that the controller responds to unknown external perturbations to a satisfying degree likely due to the efficient learning of MADQN and the capability of the shielding mechanism.} Fig.~\ref{ex_5}(b), shows how the tracking error remained in the sub-millimeter zone despite the 2N disturbance. Furthermore, this proves the stability of the proposed controller under unexpected external perturbation. 
\subsection{Evaluation with a miniature Nitinol-based continuum robot}
\begin{figure}[!ht]
	\centering
	\includegraphics[scale=0.5]{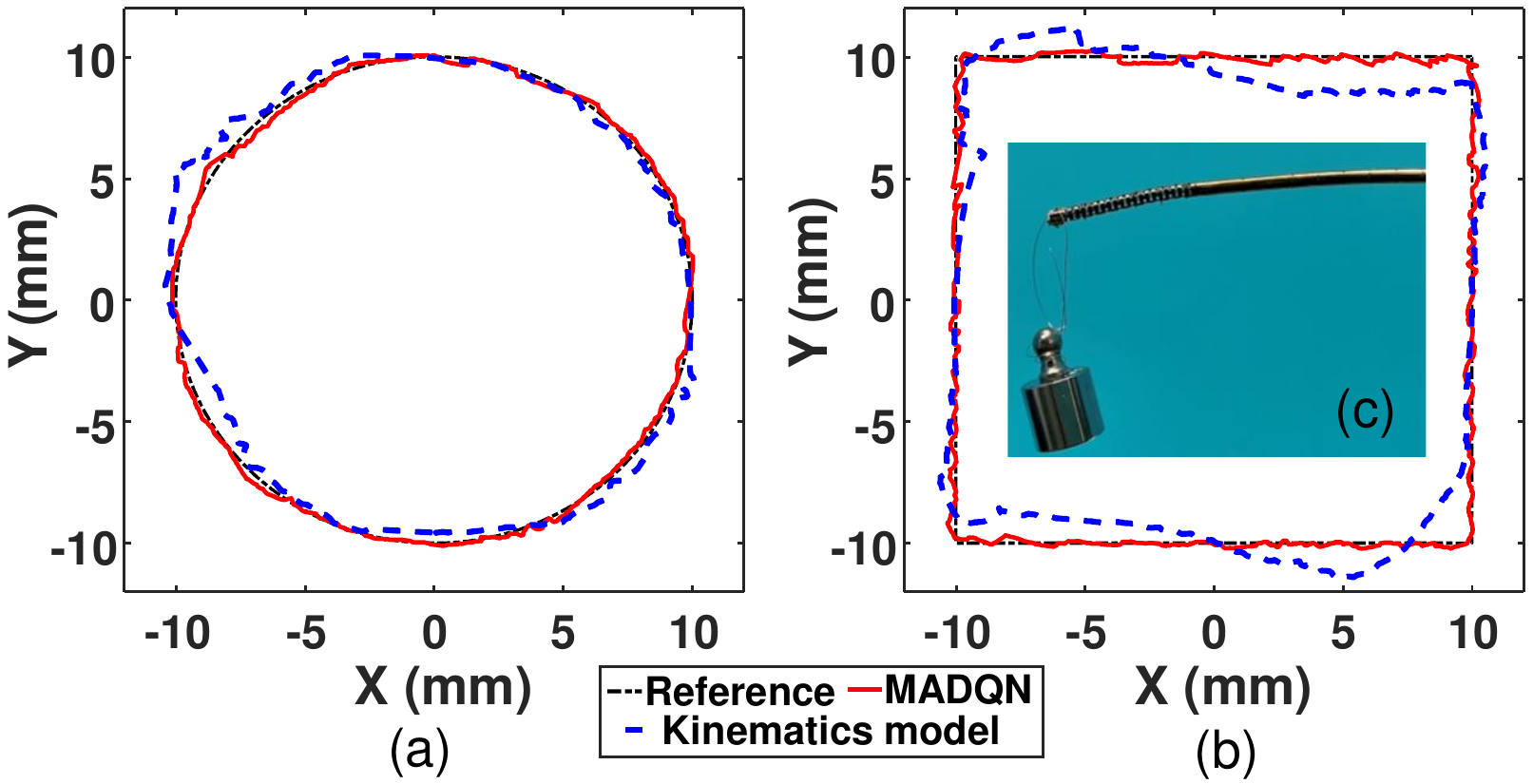}
	\caption{(a) Circular trajectory and (b) square trajectory tracking results under a 10~g external payload of a miniature Nitinol-based continuum robot (shown in inset (c)).}
	\label{ex_h}
\end{figure}
{\color{black}
In order to demonstrate the extensibility of our algorithm to other continuum robots with different physical characteristics, MADQN with shielding was used to control a 2-DoF Nitinol notched tube-based neurosurgical robot during trajectory tracking under an external payload. The robot has a smaller outer diameter of around 2.2~mm and highly nonlinear superelastic material property~\cite{zhang2021miniature}), compared to the 3D-printed nylon robot used in the other experiments. A 10~g weight was hung at the tip of the robot (weight = 9g), providing a significant payload-to-weight ratio of 1.11. KMOC was also implemented for comparison purpose. As shown in Fig.~\ref{ex_h}, MADQN achieved tracking accuracy with significantly lower RMS errors compared to the kinematics-based controller (circle: 0.41~ mm and square: 0.45~mm vs. circle: 1.22~mm, square: 2.14~mm). The superior performance of MADQN over the kinematics-based controller was even more significant than that demonstrated in the 3D printed robot in Fig.~\ref{ex_3b}. This suggests that a data-driven control algorithm such as MADQN could be more capable than a simplified kinematics-based model to adapt to unexpected external disturbances and complex internal nonlinearities to provide safe and  stable position control . This further justifies the potential of the application of the proposed shielded MADQN in continuum surgical manipulators. 
}
\section{Conclusion}
In this paper we presented the superiority of controlling a cable-driven continuum robot using a multi-agent DQN with shielding compared to traditional kinematic-based optimal controllers. 
Our simulation and physical experiments demonstrated that MADQN consistently yielded submillimeter tracking accuracy and high stability when tested under different trajectories, external payloads, soft and hard disturbances, as well as in a miniature robot with high inherent structural nonlinearity. 
Future work will involve testing the algorithm with a kinematically redundant continuum robot with significant correlation among its DoFs. Contact force will be considered as a state parameter in the shielding scheme for compliant control on the surgical manipulator.
\ifCLASSOPTIONcaptionsoff
  \newpage
\fi
\bibliographystyle{unsrt}
\bibliography{references}
\end{document}

%% file: math/rap2texdefs.tex
\usepackage{amsmath,amsfonts,amssymb}

\newcommand{\beq}{\begin{equation}}
\newcommand{\eeq}{\end{equation}}
\newcommand{\bear}{\begin{eqnarray}}
\newcommand{\bears}{\begin{eqnarray*}}
\newcommand{\eear}{\end{eqnarray}}
\newcommand{\eears}{\end{eqnarray*}}
\newcommand{\bdm}{\begin{displaymath}}
\newcommand{\edm}{\end{displaymath}}
\newcommand{\lba}{\left[\begin{array}}
\newcommand{\ear}{\end{array}\right]}

%% file: main.bbl
\begin{thebibliography}{10}

\bibitem{burgner2015continuum}
Jessica Burgner-Kahrs, D~Caleb Rucker, and Howie Choset.
\newblock Continuum robots for medical applications: A survey.
\newblock {\em IEEE Transactions on Robotics}, 31(6):1261--1280, 2015.

\bibitem{chikhaoui2018control}
MT~Chikhaoui and J~Burgner-Kahrs.
\newblock Control of continuum robots for medical applications: State of the
  art.
\newblock In {\em ACTUATOR 2018; 16th International Conference on New
  Actuators}, pages 1--11. VDE, 2018.

\bibitem{webster2010design}
Robert~J Webster~III and Bryan~A Jones.
\newblock Design and kinematic modeling of constant curvature continuum robots:
  A review.
\newblock {\em The International Journal of Robotics Research},
  29(13):1661--1683, 2010.

\bibitem{alqumsan2019robust}
Ahmad~Abu Alqumsan, Suiyang Khoo, and Michael Norton.
\newblock Robust control of continuum robots using cosserat rod theory.
\newblock {\em Mechanism and Machine Theory}, 131:48--61, 2019.

\bibitem{bieze2018finite}
Thor~Morales Bieze, Frederick Largilliere, Alexandre Kruszewski, Zhongkai
  Zhang, Rochdi Merzouki, and Christian Duriez.
\newblock Finite element method-based kinematics and closed-loop control of
  soft, continuum manipulators.
\newblock {\em Soft {Robotics}}, 5(3):348--364, 2018.

\bibitem{xu2017data}
Wenjun Xu, Jie Chen, Henry~YK Lau, and Hongliang Ren.
\newblock Data-driven methods towards learning the highly nonlinear inverse
  kinematics of tendon-driven surgical manipulators.
\newblock {\em The International Journal of Medical Robotics and Computer
  Assisted Surgery}, 13(3):e1774, 2017.

\bibitem{yip2014model}
Michael~C Yip and David~B Camarillo.
\newblock Model-less feedback control of continuum manipulators in constrained
  environments.
\newblock {\em IEEE Transactions on Robotics}, 30(4):880--889, 2014.

\bibitem{li2017model}
Minhan Li, Rongjie Kang, David~T Branson, and Jian~S Dai.
\newblock Model-free control for continuum robots based on an adaptive kalman
  filter.
\newblock {\em IEEE/ASME Transactions on Mechatronics}, 23(1):286--297, 2017.

\bibitem{lee2017nonparametric}
Kit-Hang Lee, Denny~KC Fu, Martin~CW Leong, Marco Chow, Hing-Choi Fu, Kaspar
  Althoefer, Kam~Yim Sze, Chung-Kwong Yeung, and Ka-Wai Kwok.
\newblock Nonparametric online learning control for soft continuum robot: An
  enabling technique for effective endoscopic navigation.
\newblock {\em Soft {Robotics}}, 4(4):324--337, 2017.

\bibitem{wang2021hybrid}
Ziwen Wang, Teng Wang, Baoliang Zhao, Yucheng He, Ying Hu, Bing Li, Peng Zhang,
  and Max Q-H Meng.
\newblock Hybrid adaptive control strategy for continuum surgical robot under
  external load.
\newblock {\em IEEE Robotics and Automation Letters}, 6(2):1407--1414, 2021.

\bibitem{thuruthel2018model}
Thomas~George Thuruthel, Egidio Falotico, Federico Renda, and Cecilia Laschi.
\newblock Model-based reinforcement learning for closed-loop dynamic control of
  soft robotic manipulators.
\newblock {\em IEEE Transactions on Robotics}, 35(1):124--134, 2018.

\bibitem{liu2020distance}
Wuji Liu, Zhongliang Jing, Han Pan, Lingfeng Qiao, Henry Leung, and Wujun Chen.
\newblock Distance-directed target searching for a deep visual servo sma driven
  soft robot using reinforcement learning.
\newblock {\em Journal of Bionic Engineering}, 17(6):1126--1138, 2020.

\bibitem{you2017model}
Xuanke You, Yixiao Zhang, Xiaotong Chen, Xinghua Liu, Zhanchi Wang, Hao Jiang,
  and Xiaoping Chen.
\newblock Model-free control for soft manipulators based on reinforcement
  learning.
\newblock In {\em 2017 IEEE/RSJ International Conference on Intelligent Robots
  and Systems (IROS)}, pages 2909--2915. IEEE, 2017.

\bibitem{jiang2021hierarchical}
Hao Jiang, Zhanchi Wang, Yusong Jin, Xiaotong Chen, Peijin Li, Yinghao Gan, Sen
  Lin, and Xiaoping Chen.
\newblock Hierarchical control of soft manipulators towards unstructured
  interactions.
\newblock {\em The International Journal of Robotics Research}, page
  0278364920979367, 2021.

\bibitem{mnih2013playing}
Volodymyr Mnih, Koray Kavukcuoglu, David Silver, Alex Graves, Ioannis
  Antonoglou, Daan Wierstra, and Martin Riedmiller.
\newblock Playing atari with deep reinforcement learning.
\newblock {\em arXiv preprint arXiv:1312.5602}, 2013.

\bibitem{lillicrap2015continuous}
Timothy~P Lillicrap, Jonathan~J Hunt, Alexander Pritzel, Nicolas Heess, Tom
  Erez, Yuval Tassa, David Silver, and Daan Wierstra.
\newblock Continuous control with deep reinforcement learning.
\newblock {\em arXiv preprint arXiv:1509.02971}, 2015.

\bibitem{satheeshbabu2019open}
Sreeshankar Satheeshbabu, Naveen~Kumar Uppalapati, Girish Chowdhary, and Girish
  Krishnan.
\newblock Open loop position control of soft continuum arm using deep
  reinforcement learning.
\newblock In {\em 2019 International Conference on Robotics and Automation
  (ICRA)}, pages 5133--5139, 2019.

\bibitem{satheeshbabu2020continuous}
Sreeshankar Satheeshbabu, Naveen~K Uppalapati, Tianshi Fu, and Girish Krishnan.
\newblock Continuous control of a soft continuum arm using deep reinforcement
  learning.
\newblock In {\em 2020 3rd IEEE International Conference on Soft Robotics
  (RoboSoft)}, pages 497--503. IEEE, 2020.

\bibitem{bucsoniu2010multi}
Lucian Bu{\c{s}}oniu, Robert Babu{\v{s}}ka, and Bart De~Schutter.
\newblock Multi-agent reinforcement learning: An overview.
\newblock {\em Innovations in multi-agent systems and applications-1}, pages
  183--221, 2010.

\bibitem{ansari2016multiagent}
Yasmin Ansari, Egidio Falotico, Yoan Mollard, Baptiste Busch, Matteo
  Cianchetti, and Cecilia Laschi.
\newblock A multiagent reinforcement learning approach for inverse kinematics
  of high dimensional manipulators with precision positioning.
\newblock In {\em 2016 6th IEEE International Conference on Biomedical Robotics
  and Biomechatronics (BioRob)}, pages 457--463. IEEE, 2016.

\bibitem{perrusquia2020multi}
Adolfo Perrusqu{\'\i}a, Wen Yu, and Xiaoou Li.
\newblock Multi-agent reinforcement learning for redundant robot control in
  task-space.
\newblock {\em International Journal of Machine Learning and Cybernetics},
  pages 1--11, 2020.

\bibitem{ansari2017multiobjective}
Yasmin Ansari, Mariangela Manti, Egidio Falotico, Matteo Cianchetti, and
  Cecilia Laschi.
\newblock Multiobjective optimization for stiffness and position control in a
  soft robot arm module.
\newblock {\em IEEE Robotics and Automation Letters}, 3(1):108--115, 2017.

\bibitem{diallo2017learning}
Elhadji Amadou~Oury Diallo, Ayumi Sugiyama, and Toshiharu Sugawara.
\newblock Learning to coordinate with deep reinforcement learning in doubles
  pong game.
\newblock In {\em 2017 16th IEEE International Conference on Machine Learning
  and Applications (ICMLA)}, pages 14--19. IEEE, 2017.

\bibitem{alshiekh2018safe}
Mohammed Alshiekh, Roderick Bloem, R{\"u}diger Ehlers, Bettina K{\"o}nighofer,
  Scott Niekum, and Ufuk Topcu.
\newblock Safe reinforcement learning via shielding.
\newblock In {\em Proceedings of the AAAI Conference on Artificial
  Intelligence}, volume~32, 2018.

\bibitem{ding2021towards}
Qingpeng Ding, Yongkang Lu, and Shing~Shin Cheng.
\newblock Towards a multi-imager compatible continuum robot under real-time
  performance of modular {SMA}.
\newblock In {\em 2021 IEEE International Conference on Robotics and Automation
  (ICRA)}. IEEE (Accepted), 2021.

\bibitem{sutton2018reinforcement}
Richard~S Sutton and Andrew~G Barto.
\newblock {\em Reinforcement learning: An introduction}.
\newblock MIT press, 2018.

\bibitem{sefati2020surgical}
Shahriar Sefati, Rachel Hegeman, Farshid Alambeigi, Iulian Iordachita, Peter
  Kazanzides, Harpal Khanuja, Russell Taylor, and Mehran Armand.
\newblock A surgical robotic system for treatment of pelvic osteolysis using an
  fbg-equipped continuum manipulator and flexible instruments.
\newblock {\em IEEE/ASME Transactions on Mechatronics}, 2020.

\bibitem{shi2016shape}
Chaoyang Shi, Xiongbiao Luo, Peng Qi, Tianliang Li, Shuang Song, Zoran
  Najdovski, Toshio Fukuda, and Hongliang Ren.
\newblock Shape sensing techniques for continuum robots in minimally invasive
  surgery: A survey.
\newblock {\em IEEE Transactions on Biomedical Engineering}, 64(8):1665--1678,
  2016.

\bibitem{tang2012ergonomics}
Benjie Tang, Sichuan Hou, and Sir~Alfred Cuschieri.
\newblock Ergonomics of and technologies for single-port lapaxroscopic surgery.
\newblock {\em Minimally Invasive Therapy \& Allied Technologies},
  21(1):46--54, 2012.

\bibitem{zhang2021miniature}
Tianci Zhang, Zhongyuan Ping, and Siyang Zuo.
\newblock Miniature continuum manipulator with three degrees-of-freedom force
  sensing for retinal microsurgery.
\newblock {\em Journal of Mechanisms and Robotics}, 13(4):041002, 2021.

\end{thebibliography}
